\RequirePackage{fix-cm}
\documentclass[twocolumn,natbib]{svjour3}          % twocolumn
\smartqed  % flush right qed marks, e.g. at end of proof
\usepackage[T1]{fontenc}

\usepackage{graphics,graphicx}
\usepackage{url}
\usepackage{amsmath,amssymb}
\usepackage{thmtools}
\usepackage{mathtools}

\usepackage[breaklinks,hidelinks]{hyperref}  
\usepackage[hyphenbreaks]{breakurl}

\usepackage{framed, times}
\newenvironment{notework}[1]{
\vspace{1em}
\begin{leftbar}
\par
\begingroup
\noindent
\textbf{#1}:\\
\indent
}{
\endgroup
\end{leftbar}
\vspace{1em}
}

\newcommand{\tyq}[1]
{
\begin{notework}{Comment --- yuqing}
#1
\end{notework}
}

\newcommand{\eg}{e.g.}
\newcommand{\ie}{i.e.}
\newcommand{\viz}{viz.}
\newcommand{\cf}{cf.}
\newcommand{\st}{\ensuremath{\mbox{s.t.}}}

\newcommand{\set}[1]{\ensuremath{\{#1\}}}

\newcommand{\tuple}[1]{\ensuremath{\langle #1 \rangle}}

\newcommand{\setagents}{\ensuremath{\mathcal{A}}}
\newcommand{\setcommlinks}{\ensuremath{\mathcal{C}}}
\newcommand{\producer}{\ensuremath{ag}}
\newcommand{\destination}{\ensuremath{Tg}}

\newcommand{\fram}{FCA}
\newcommand{\prblm}{\mcP}
\newcommand{\M}{\mc{M}} 

\newcommand{\agent}[1]{\ensuremath{ag_{#1}}}
\newcommand{\ag}[1]{\ensuremath{\agent{#1}}}
\newcommand{\agentone}{\agent{1}}
\newcommand{\agenttwo}{\agent{2}}
\newcommand{\agentthree}{\agent{3}}
\newcommand{\agentfour}{\agent{4}}
\newcommand{\agentx}{\agent{x}}
\newcommand{\agenty}{\agent{y}}
\newcommand{\agentq}{\agent{q}}

\newcommand{\disclosuredegree}[2]{\ensuremath{x_{#1,#2}}}
\newcommand{\disclosurefun}{\ensuremath{d}}

\newcommand{\probDisclosuredegree}[2]{\ensuremath{{x_{#1,#2}}}}
\newcommand{\probSend}[2]{\ensuremath{{s_{#1,#2}}}}

\newcommand{\probInfer}{\ensuremath{y}}
\newcommand{\probInferred}[2]{\ensuremath{{y_{{#1}_{|#2}}}}}

\newcommand{\rvInference}[2]{\ensuremath{{I_{#1}({#2})}}}
\newcommand{\famRvInference}[2]{\ensuremath{{\mathcal{I}_{{#1}_{|#2}}}}}
\newcommand{\distributionInference}[2]{\ensuremath{{F_{I_{#1}}(\cdot; #2)}}}
\newcommand{\densityInference}[2]{\ensuremath{{f_{I_{#1}}(\cdot; #2)}}}

\newcommand{\rvImpact}[2]{\ensuremath{{Z_{#1}({#2})}}}

\newcommand{\distributionImpact}[2]{\ensuremath{{F_{Z_{#1}}(\cdot; #2)}}}
\newcommand{\densityImpact}[2]{\ensuremath{{f_{Z_{#1}}(\cdot; #2)}}}

\newcommand{\probImpact}{\ensuremath{z}}

\newcommand{\pz}{\widetilde{z}}
\newcommand{\pb}{\widetilde{b}}
\newcommand{\pr}{\widetilde{r}}

\newcommand{\rvBenefit}[2]{\ensuremath{{B_{#1}({#2})}}}

\newcommand{\distributionBenefit}[2]{\ensuremath{{F_{B_{#1}}(\cdot; #2)}}}
\newcommand{\densityBenefit}[2]{\ensuremath{{f_{B_{#1}}(\cdot; #2)}}}

\newcommand{\probBenefit}{\ensuremath{b}}

\newcommand{\rvRisk}[2]{\ensuremath{{R_{#1}({#2})}}}

\newcommand{\distributionRisk}[2]{\ensuremath{{F_{R_{#1}}(\cdot; #2)}}}
\newcommand{\densityRisk}[2]{\ensuremath{{f_{R_{#1}}(\cdot; #2)}}}

\newcommand{\probRisk}{\ensuremath{r}}

\newcommand{\dotop}{\ensuremath{\odot}}
\newcommand{\sumop}{\ensuremath{\oplus}}

\newcommand{\tx}[1]{\textrm{#1}}

\newcommand{\mbf}[1]{\mathbf{#1}}
\newcommand{\msf}[1]{\mathsf{#1}}
\newcommand{\mbb}[1]{\mathbb{#1}}
\newcommand{\mc}[1]{\mathcal{#1}}
\newcommand{\mcL}{\mathcal{L}}
\newcommand{\mcI}{\mathcal{I}}
\newcommand{\mcP}{\mathcal{P}}

\newcommand{\mcM}{\mathcal{M}}
\newcommand{\mcR}{\mathcal{R}}
\newcommand{\mcU}{\mathcal{U}}

\newcommand{\mcB}{\mathcal{B}}
\newcommand{\mcZ}{\mathcal{Z}}

\newcommand{\ags}{\msf{AGS}}

\begin{document}

\title{Reasoning about the Impacts of Information Sharing}
%\title{A Framework for Using Trust to Assess Risk in Information Sharing




\author{
Chatschik Bisdikian \thanks{
This research was sponsored by US Army Research laboratory and the UK Ministry of Defence and was accomplished under Agreement Number W911NF-06-3-0001. The views and conclusions contained in this document are those of the authors and should not be interpreted as representing the official policies, either expressed or implied, of the US Army Research Laboratory, the U.S. Government, the UK Ministry of Defence, or the UK Government. The US and UK Governments are authorised to reproduce and distribute reprints for Government purposes notwithstanding any copyright notation hereon.}
\and Federico Cerutti  \and Yuqing Tang  \and Nir Oren
}


\institute{ Chatschik Bisdikian \at
  IBM Research Division,Thomas J. Watson Research Center,P.O. Box 704,Yorktown Heights, NY 10598, USA\\
  \email{bisdik@us.ibm.com} \and Federico Cerutti \at
  University of Aberdeen, School of Natural and Computing Science, King's College, \\ AB24 3UE, Aberdeen, UK\\
  \email{f.cerutti@abdn.ac.uk} \and Yuqing Tang \at
  Carnegie Mellon University, Robotics Institute, 5000 Forbes Ave, Pittsburgh, PA 15213, USA\\
  \email{yuqing.tang@cs.cmu.edu} \and Nir Oren \at
  University of Aberdeen, School of Natural and Computing Science, King's College, \\ AB24 3UE, Aberdeen, UK\\
  \email{ n.oren@abdn.ac.uk} }

\date{Received: date / Accepted: date}

\maketitle

\begin{abstract}
  In this paper we describe a decision process framework allowing an
  agent to decide what information it should reveal to its neighbours
  within a communication graph in order to maximise its utility. We
  assume that these neighbours can pass information onto others within
  the graph. The inferences made by agents receiving the messages can have a positive or negative impact on the information providing agent, and our decision process seeks to identify how a message should be modified in order to be most beneficial to the information producer. Our decision process is based on the provider's subjective beliefs about others in the system, and therefore makes extensive use of the notion of trust. Our core contributions are therefore the construction of a model of information propagation; the description of the agent's decision procedure; and an analysis of some of its properties.
\end{abstract}

\section{Introduction}
Appropriate decision making by an agent operating within a multi-agent system often requires information from other agents. However, unless the system is fully cooperative, there are typically both costs and benefits to divulging information --- while the agent may be able to achieve some goals, others might be able to use this information to their advantage later. An agent must therefore weigh up the costs and benefits that information divulgence will bring it when deciding how to act. One of the most critical factors in this calculation is the trust placed in the entity to which one is providing the information --- an untrusted individual might pass private information onto others, or may act upon the information in a manner harmful to the information provider.

In this paper we seek to provide a trust based decision mechanism for assessing the positive and negative effects of information release to an agent. Using our mechanism, first discussed in \cite{Bisdikian2013} and expanded here, the agent can decide how much information to provide in order to maximise its own utility. We situate our work within the context of a multi-agent system. Here, an agent must assess the risk of divulging information to a set of other agents, who in turn may further propagate the information. The problem the agent faces is to identify the set of information that must be revealed to its neighbours (who will potentially propagate the information further) in order to maximise its own utility. 

In the context of a multi-agent system, the ability of an agent to assess the risk of information sharing is critical when agents have to reach agreement, for example when coordinating, negotiating or delegating activities. In many contexts, agents have conflicting goals, and inter-agent interactions must take the risk of a hidden agenda into account. Thus, a theory of risk assessment for determining the right level of disclosure to apply to shared information is vital in order to avoid undesirable impacts on an information producer.

As a concrete example, consider the work described in  \cite{Chakraborty2012}, where information from accelerometer data attached to a person can be used to make either \emph{white-listed} inferences --- that the person desires others to infer, or \emph{black-listed} inferences --- which the person would rather not reveal. For example, the person may wish a doctor to be able to determine how many calories they burn in a day, but might not want others to be able to infer  their state (e.g. sitting, running or asleep). The person must thus identify which parts of the accelerometer data should be shared in order to enable or prevent their white- or black-listed inferences. While \cite{Chakraborty2012} examined how inferences can be made (e.g. that the sharing of the entropy of FFT coefficients provides a high probability of detecting activity level and low probability of detecting activity type), this work did not consider the \emph{impacts} of sharing such information when it is passed on to others.

In this paper we focus on the case where we assume that black- and white-listed inferences can be made by other agents within a system, and seek to identify what information to provide in order to obtain the best possible outcome for the information provider. 

%We present a running example, used throughout the paper, which illustrates such a scenario.
To illustrate such a scenario, let us consider a governmental espionage agency which has successfully placed spies within some hostile country. It must communicate with these spies through a series of handlers, some of which may turn out to be double-agents. It must therefore choose what information to reveal to these handlers in order to maximise the benefits that spying can bring to it, while minimising the damage they can do. It is clear that the choices made by the agency depend on several factors. First, it must consider the amount of trust it places in the individual spies and handlers. Second, it must also take into account the amount of harm these can do with any information it provides to them. Finally, it must consider the benefits that can accrue from providing its spies with information. The combination of the first and second factors together provide a measure of the negative effects of information sharing. Now when considering the second factor, an additional detail must be taken into account, namely that the information recipients (i.e. the spies) may already have some knowledge which, when combined with the information provided by the agency, will result in additional unexpected information being inferred. Therefore, the final level of harm which the agency may face depends not  on the information it provides, but instead on the undesired inferences which hostile spies can make.

The remainder of this paper is structured as follows. In Section \ref{sec:model-risk-inform} we describe our model, outlining the process of decision making that an agent performs in the presence of white- and black-listed inferences. We concentrate on a special case of communication in multi-agent systems, and show how such a case can be reduced to communication between an information provider and consumer (Section \ref{sec:communicationNetworks}). We describe the decision procedure in Section \ref{sec:decisionProcess}.  Section \ref{sec:scenario} provides a numeric example of the functioning of our system. We then contrast our approach with existing work in Section \ref{sec:disc-future-work}, and identify several avenues of future work. Section \ref{sec:concl-future-works} concludes the paper. Appendix \ref{sec:case-cont-rand} discusses the relevant properties of this approach when considering continuous random variables: in the following the will mainly focus on the case of discrete random variables.

%\section{Background}
%\label{sec:background}

\section{The Effects of Information Sharing}
\label{sec:model-risk-inform}

We consider a situation where an information producer shares information with one or more information consumers.  These consumers can, in turn, forward the information to others, who may also forward it on, repeating the cycle. Furthermore, since a consumer may or may not use the information provided as expected by the provider, the producer must assess the damage it will incur if the provided information is misused. The decision problem faced by the producer is to therefore identify an appropriate message to send to a consumer which will achieve an appropriate balance between desired and undesired effects. We assume that once a information is provided, the producer is unable to control its spread or use further .

We begin by describing a model of such a system. As part of our
notation, we use upper-case letters, \eg{} $X$, to represent random
variables (r.v.'s); lower-case letters, \eg{} $x$, to represent
realisation instances of them, and $F_X (\cdot)$ and $f_X (\cdot) $ to
represent the probability distribution and density of the r.v. $X$,
respectively; $Pr(\cdot)$ and $Pr(\cdot \mid \cdot)$ to represent the
probability and conditional probability of discrete random variables
respectively.

%\subsection{The Model}

We consider a set of agents able to interact with their neighbours
through a set of communication links, as embodied by a communication
graph or network. We assume that each agent knows the topology of this
network. We introduce the concept of a \emph{Framework for Communication
  Assessment} \fram{} that considers the set of agents, the messages that can be exchanged, the
communication links of each agents, a producer that is willing to
share some information, and the recipients of the information, which are
directly connected to the producer within the communication graph.

\begin{definition}
  A \emph{Framework for Communication Assessment} (\fram) is a 5-ple:
  \[
  \tuple{\setagents, \setcommlinks, \M, \producer, m}
  \]
 where:
 \begin{itemize}
 \item $\setagents$ is a set of agents;
 \item $\setcommlinks \subseteq \setagents \times \setagents$ is the
   set of communication links among agents;
 \item $\M$ is the set of all the messages that can be exchanged;
 \item $\producer \in \setagents$ is the \emph{producer}, \viz{} the
   agent that shares information;
 \item $m \in \M$ is a message that is sent by the producer $\producer$
   and whose impact is being assessed;
 \item $\setagents \setminus \set{\producer}$ is the set of
   \emph{consumers}.
% , and in particular:
   % \begin{itemize}
   % \item $\destination \subseteq \setagents \setminus \set{\producer}$
   %   are the \emph{desired consumers}, and $\forall \agentx \in
   %   \destination, \tuple{\producer, \agentx} \in \setcommlinks$;
   % \item $\setagents \setminus (\set{\producer} \cup \destination)$,
   %   are the \emph{undesired consumers}.
   % \end{itemize}

   % \item $\forall \agentx \in \destination, \tuple{\producer,
   %   \agentx}
   %   \in \setcommlinks$;
   % \item $\forall \agentx \in \destination$, $\agentx$ is a
   %   \emph{desired consumer} for the message $m$;
   % \item $\forall \agentx \in \setagents \setminus (\set{\producer}
   %   \cup \destination)$, $\agentx$ is an \emph{undesired
   %   consumer}.
 \end{itemize}
\end{definition}
%

% \begin{definition}
%   A \emph{Framework for Risk Assessment} (\fram) is a triple:

%   \[
%   \fram = \tuple{\setagents, \setcommlinks, \M}
%   \]
%   \noindent
%   where:
%   \begin{itemize}
%   \item $\setagents$ is a set of agents;
%   \item $\setcommlinks \subseteq \setagents \times \setagents$ is the
%     set of communication links among agents;
%   \item $M$ is a set of all the messages that can be exchanged;
%   \end{itemize}
%   Given a risk assessment framework $\fram$, a problem instance of
%   risk assessment is a triple:
%   \[
%   \prblm = \tuple{\producer, m, \destination}
%   \]
%   where
%   \begin{itemize}
%   \item $\producer \in \setagents$ is a \emph{producer}, \viz{} the
%     agent that shares information with a set of consumers
%     $\destination \subseteq \setagents$.
%   \item $m \in M$ is a message that is being assessed;
%   \item $\destination$ is a set of \emph{desired consumer}s for the
%     message such that
%     \begin{itemize}
%     \item $\forall \agentx \in \destination, \tuple{\producer,
%         \agentx} \in \setcommlinks$;
%     \item $\forall \agentx \in \setagents \setminus (\set{\producer}
%       \cup \destination)$, $\agentx$ is an \emph{undesired consumer}.
%     \end{itemize}
%   \end{itemize}
% \end{definition}

Given a framework $\fram$, \producer{} will make use of the procedure
described in this paper to determine how to share information. This
information sharing decision seeks to identify a \emph{degree of
  disclosure} for the original message - reducing the information
provided according to this degree of disclosure results in a
\emph{derived} version of the original message, which (informally)
conveys less information than the original.
% %
% \begin{definition}
%   Given a \fram{} $\tuple{\setagents, \setcommlinks, \M, \producer, m, \destination}$, $\forall \agentone, \agenttwo \in \setagents \setminus \set{\producer}$:
%   \begin{itemize}
%   \item $\probSend{\agentone}{\agenttwo} \in [0,1]$ is the probability
%     that $\agentone$ will propagate to $\agenttwo$ the disclosed part
%     of $m$ that it receives;
%   \item $\probDisclosuredegree{\agentone}{\agenttwo} \in [0,1]$ is the
%     assumed disclosure degree of communications between the two
%     agents.
%   \end{itemize}
% \end{definition}
% %
As an example, if the agency knows that ``country A is going to invade
country B'', and is deciding whether to communicate this, a message of
the form ``country B is going to be invaded'' is a derived message,
obtained when the degree of disclosure is less than 1.  We do not
specify the exact mapping and the corresponding mathematical
properties between the degree of disclosure and the derived message in
this work, leaving this as a future avenue of research.
\begin{definition}
  Given $\setagents$ a set of agents, a message $m \in \M$, $\ag{i},
  \ag{j} \in \setagents$, $\disclosuredegree{i}{j}(m) \in [0,1]$ is
  the \emph{degree of disclosure} by which agent \ag{i} will send the
  message $m$ to agent \ag{j}, where $\disclosuredegree{i}{j}(m)=0$
  implies no sharing and $\disclosuredegree{i}{j}(m)=1$ implies full
  disclosure between the two agents.  We define the \emph{disclosure
    function} as follows:
  \[
  \disclosurefun: \M \times [0,1] \mapsto \M
  \]
  $\disclosurefun(\cdot,\cdot)$ accepts a message and a degree of
  disclosure (for that message) as its inputs, and returns a modified
  message (referred to as the disclosed portion of the original
  message). 
\end{definition}
%
%% Yuqing: We don't need to say the following at this point. It will confuse readers. 
% Since we concentrate on a single message, we will where evident from
% the context omit the source message to which the degree of
% disclosure is related.
In this paper, we consider a specific version of  degree of
disclosure, interpreting $\disclosuredegree{i}{j}(m)$ as a conditional
probability by which agent $\ag{i}$ will modify message $m$ into a new
message $m'$ when communicating with agent $\ag{j}$:
\begin{align*}
\disclosuredegree{i}{j}(m) &= Pr_{i, j}(m' \mid m ) 
\end{align*}
where $m' \in \M$ is a modified message of $m$.  Note that in this
interpretation $ m ' = \disclosurefun(m, \disclosuredegree{i}{j}(m))$
(if there are multiple messages with the same degree of disclosure
relative to $m$, the disclosure function $\disclosurefun$ will select
one among them randomly).
% The following sentence is to justify the monotonicity assumption in
% the communication network. As the monotonicity requires measuring
% the relative content level of messages.
 A more sophisticated notation of message
disclosure level and its probabilistic interpretation, such as
separating the models of measuring the relative content level of
messages and the corresponding conditional probabilities of message
transformation during  communication, will be left for our future
work.
  
%  we obtain a
% \emph{message disclosure policy} for $\agentone$ when communicating
% with $\agenttwo$:
% \begin{align*}
% \disclosuredegree{\agentone}{\agenttwo}(m) &= Pr_{\agentone, \agenttwo}(m' \mid m ) 
% \end{align*}
% That is,
% $Pr_{\agentone, \agenttwo}(\cdot \mid \cdot )$ describes the likelihood that on receiving a message $m$, agent $\agentone$ will transmit the message $m'$ to agent $\agenttwo$.

Given a \fram, the decision whether or not to share the information
with the recipient must consider the impact that the information
recipient can incur to the producer.  We assume that agents within the
system are selfish --- an information provider will only share
information if doing so provides it with some benefit but with the
damage being as less as possible. However, such a benefit and damage
may be uncertain. Therefore, when sharing information, the producer
not only considers the benefit it obtains, but must also considers
potential negative side effects based on the following:

\begin{enumerate}
\item the probability that an agent in possession of the message will forward it onward;
\item the levels of disclosure of messages exchanged between two agents;
\item the ability of each agent to infer knowledge from the received (disclosed) message;
\item the \emph{impacts} (i.e. positive and negative effects) that the inferred knowledge has on the information producer.
\end{enumerate}

The following definition therefore models the uncertainty of the impact of sharing information via a random variable.

\begin{definition}
  \label{def:rv-impact}
  Given a \fram{} $\tuple{\setagents, \setcommlinks, \M, \producer, m}$, let $\ag{i} \in \setagents \setminus \set{\producer}$,
  $Z(\disclosuredegree{\producer}{i}) \in \mbf{Z}$ be a
  r.v. which represents the \emph{impact} agent \producer{} receives
  when sharing the message $m$ with a degree of disclosure
  $\disclosuredegree{\producer}{i}$ with agent \ag{i}. $\mbf{Z}$ is called 
  the \emph{space  of impact}. $Z(\disclosuredegree{\producer}{i})$ can either be 
  \begin{itemize}
  \item a continuous random variable whose distribution is described
    by $F_Z (\cdot ; \disclosuredegree{\producer}{i})$ and $f_Z
    (\cdot ; \disclosuredegree{\producer}{i})$, or
  \item a discrete random variable whose probability is $Pr( z \mid
    \disclosuredegree{\producer}{i})$ where $z =
      Z(\disclosuredegree{\producer}{i})$.
    \end{itemize}
\end{definition}

The central theme of this paper is centred around
Definition~\ref{def:rv-impact}. More specifically, we focus on 1) how to derive the distribution of
impact from the disclosure degree of messages; 2) how to evaluate the positive and negative effects
 of the impact; and 3) how to make decisions regarding the disclosure degree of messages based on impact.

\section{Communication Networks}\label{sec:communicationNetworks}
We now turn our attention to communication between agents who must send information via an intermediary.  We show that under some special conditions, such communication can be abstracted as direct communication utilising a different degree of disclosure. In order to show this result, we introduce two operators combining disclosure degrees when information is shared in this way. The first operator \emph{discounts} the degree of disclosure based on agents within the message path, while the second operator \emph{fuses} information which may have travelled along multiple paths.

\begin{figure}[!htb]
  \centering
  \includegraphics[width=0.4\textwidth]{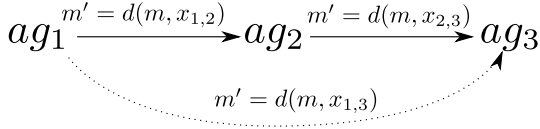}
  \caption{Combining disclosure degrees: single communication path
    (Definition~\ref{def:dotop}).\label{fig:def-dotop}}
\end{figure}

Figure \ref{fig:def-dotop} depicts the simplest case where the first
operator can be applied. Let us suppose that there are three agents
$\ag{1}, \ag{2}, \ag{3}$ and that $\ag{1}$ plays the role of the
producer of the information $m$, that is shared with $\ag{1}$ with a
degree of disclosure $\probDisclosuredegree{1}{2}$ (the message sent
to $\agenttwo$ is $m'=\disclosurefun(m,
\probDisclosuredegree{1}{2})$. $\ag{2}$ then shares what it received
($m'$) with $\ag{3}$, sending it a message $\disclosurefun(m',
\probDisclosuredegree{1}{3})$ derived from a new degree of disclosure
$\probDisclosuredegree{2}{3}$ which applies to $m'$ instead of $m$.
Concerning our scenario, it can be the case that the agency ($\ag{1}$)
cannot send a message directly to its agent in the hostile
country. Therefore it has to deliver the message through an
intermediate agent $\ag{2}$. Unfortunately, the agency knows that
there is a possibility that $\ag{2}$ will propagate the message
towards $\ag{3}$, who is an hostile spy.

The operator we introduce in Definition~\ref{def:dotop} computes the
equivalent disclosure degree
($\probDisclosuredegree{1}{2}$) which should be used
for deriving a new message from $m$ that $\ag{1}$ can send to
$\ag{3}$ and such that $\disclosurefun(m',
\probDisclosuredegree{2}{3}) = \disclosurefun(m,
\probDisclosuredegree{1}{3})$. We require the
introduced operator to be (i) transitive, and (ii) that the returned
value $\probDisclosuredegree{1}{3}$ should not be
greater than $\probDisclosuredegree{1}{2}$. This
``monotonicity'' requirement is built on the intuition that
$\ag{2}$ does not know the original information $m$, just its
derived version $\disclosurefun(m,
\probDisclosuredegree{1}{2})$. It also rests on the
critical assumption that $\ag{2}$ does not make any kind of
inference before sharing its knowledge. For instance, if $\ag{1}$
shares with $\ag{2}$ a message with a degree of disclosure of 0.7
of the original message $m$, and $\ag{2}$ shares with $\ag{3}$
a message with a degree of disclosure of 0.5 of the message it
received, one could argue that $\ag{3}$ a message with a degree
of disclosure of 0.35 of the original message $m$. The monotonicity
requirement is strictly related to the assumption that an agent cannot
share what it derived: this requirement will be relaxed in future
developments of the \fram{} framework.

\begin{definition}
  \label{def:dotop}
  Given a \fram{} $\tuple{\setagents, \setcommlinks, \M, \producer, m}$, for any three agents $\ag{1}, \ag{2}, \ag{3} \in
  \setagents$, $\tuple{\ag{1}, \ag{2}}$, $\tuple{\ag{2},
    \ag{3}} \in \setcommlinks$, let $m' = \disclosurefun(m,
  \probDisclosuredegree{1}{2})$ be the message sent by
  $\ag{1}$ to $\ag{2}$ (see Fig. \ref{fig:def-dotop}). Then the
  message sent by $\ag{1}$ to $\ag{3}$ is:
    \[
    \disclosurefun(m', \probDisclosuredegree{2}{3})
    = \disclosurefun(m, \probDisclosuredegree{1}{3})
    \]
    where
    \begin{itemize}
    \item $\probDisclosuredegree{1}{3} =
      \tuple{%\probSend{\agentone}{\agenttwo},
        \probDisclosuredegree{1}{2}} \dotop
      \tuple{%\probSend{\agenttwo}{\agentthree},
        \probDisclosuredegree{2}{3}}$;
    \item $\dotop$ is a transitive function such that
      \[
      \dotop: ([0,1]) \times ([0,1]) \mapsto
      [0,1]
      \]
    \item $\probDisclosuredegree{1}{3} \leq
      \probDisclosuredegree{1}{2}$.
    \end{itemize}
  \end{definition}

\begin{figure}[!htb]
  \centering
  \includegraphics[width=0.35\textwidth]{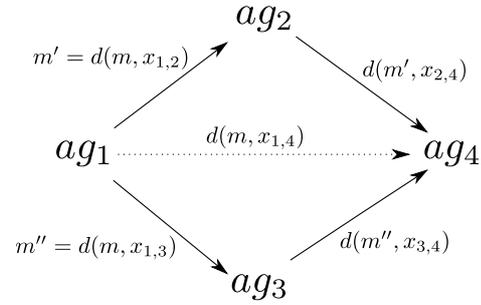}
  \caption{Combining disclosure degrees: multiple communication path (Definition~\ref{def:sumop}).\label{fig:def-sumop}}
\end{figure}

  Our second operator deals with the case where there are multiple
  path that a message can traverse before reaching an information
  consumer. This is the case depicted in Fig. \ref{fig:def-sumop},
  where $\ag{1}$ shares the same information with $\ag{2}$ and
  $\ag{3}$, but filters these using different degrees of
  disclosure $\probDisclosuredegree{1}{2}$ and
  $\probDisclosuredegree{1}{3}$ for the two agents
  respectively. Following this, $\ag{2}$ and $\ag{3}$ share
  the information they obtained with $\ag{4}$.
For instance, it can be the case that the agency, trying to reach its agent, shows information with both $\ag{2}$ and $\ag{3}$, hoping that somehow the message will eventually reach the agent. As before, the agency is aware that both $\ag{2}$ and $\ag{3}$ have contacts with an enemy spy $\ag{4}$.
 
We therefore define a transitive operator (in Definition
\ref{def:sumop} below) which when used with the operator defined in
Definition \ref{def:dotop} above, provides us with an equivalent
degree of disclosure as if the message was sent directly from
$\agentone$ to $\agentfour$
($\probDisclosuredegree{1}{4}$). This operator must
honour the monotonicity requirement --- the derived degree of
disclosure $\probDisclosuredegree{1}{4}$ cannot be
greater than the minimum of the disclosure degrees used for sharing
the information with $\agenttwo$
($\probDisclosuredegree{1}{2}$) and with $\agentthree$
($\probDisclosuredegree{1}{3}$). As for the previous
operator,we assume that intermediate agents do not make any inferences
before sharing information.

\begin{definition}
  \label{def:sumop}
  Given a \fram{} $\tuple{\setagents, \setcommlinks, \M, \producer,
    m}$, $\forall \agentone, \agenttwo, \allowbreak \agentthree, \allowbreak
  \agentfour \in \setagents$, $\tuple{\agentone, \agenttwo}$,
  $\tuple{\agentone, \agentthree}$, $\tuple{\agenttwo, \agentfour}$,
  \allowbreak $\tuple{\agentthree, \agentfour} \in \setcommlinks$, let
  $m' = \disclosurefun(m, \probDisclosuredegree{1}{2})$ be the message
  sent by $\agentone$ to $\agenttwo$, and let $m'' = \disclosurefun(m,
  \probDisclosuredegree{1}{3})$ be the message sent by $\agentone$ to
  $\agentthree$ (see Fig. \ref{fig:def-sumop}). Then there is a
  \emph{merge} function that merges the message sent by $\agenttwo$ to
  $\agentfour$, with the message sent by $\agentthree$ to $\agentfour$
  as follows:
  \[
  merge(\disclosurefun(m', \probDisclosuredegree{2}{4}),
  \disclosurefun(m'', \probDisclosuredegree{3}{4})) =
  \disclosurefun(m, \probDisclosuredegree{1}{4})
  \]
  where
  \begin{itemize}
  \item $\probDisclosuredegree{1}{4}$ is defined as: \begin{align*}
      \probDisclosuredegree{1}{4} &= \left(
        \tuple{%\probSend{\agentone}{\agenttwo},
          \probDisclosuredegree{1}{2}} \dotop \tuple{%\probSend{2}{4},
          \probDisclosuredegree{2}{4}} \right) \sumop \left(
        \tuple{%\probSend{\agentone}{\agentthree},
          \probDisclosuredegree{1}{3}} \dotop
        \tuple{%\probSend{\agentthree}{\agentfour},
          \probDisclosuredegree{3}{4}} \right);
    \end{align*}
    % $\probDisclosuredegree{\agentone}{\agentfour} = \left(
    %   \tuple{\probSend{\agentone}{\agenttwo},
    %   \probDisclosuredegree{\agentone}{\agenttwo}} \dotop
    %   \tuple{\probSend{\agenttwo}{\agentfour},
    %   \probDisclosuredegree{\agenttwo}{\agentfour}} \right)
    % \allowbreak \sumop
    % \left( \tuple{\probSend{\agentone}{\agentthree},
    %   \probDisclosuredegree{\agentone}{\agentthree}} \dotop
    %   \tuple{\probSend{\agentthree}{\agentfour},
    %   \probDisclosuredegree{\agentthree}{\agentfour}} \right)$;
  \item $\sumop$ is a transitive function \st{}
    \[
    \sumop: [0,1] \times [0,1] \mapsto [0,1]
    \]
  \item $\probDisclosuredegree{1}{4} \leq
    \min{\set{\probDisclosuredegree{1}{2},
        \probDisclosuredegree{1}{3}}}$.
  \end{itemize}
\end{definition}

  Having described the properties required of $\dotop$ and $\sumop$,
  we do not instantiate them further. However, any operators which
  satisfy these properties allow us to treat the transmission of
  information between any two agents in the network as a transmission
  between directly connected agents, subject to changes in the degree
  of disclosure. Computing this degree of disclosure requires
  $\producer$ to make some assumptions, which can be computed from
  specific $\dotop$ and $\sumop$ instantiations. In the remainder of
  the paper, we therefore consider communication only from an
  information source to the information consumer (ignoring
  intermediary agents), and deal with a single degree of disclosure
  used in this communication. This ``derived'' degree of disclosure is
  computed using a specific instantiation of $\dotop$ and $\sumop$ by
  considering the message path through the network from the
  information producer to the consumer..

  %\aq{q}: the agent in {q}uestion.
  After a message $m \in \M$ from producer $\producer$ propagates
  through the communication work, we obtain a distribution over the
  messages that a consumer $\ag{q}$ can receive: $Pr_{q}(m_i)$ for
  each $m_i \in \M$. We represent this distribution through vector
  notation:
\[
\vec{x}_{q} =    
\begin{pmatrix}
   Pr_{q}(m_1) \\
   Pr_{q}(m_2)  \\
  \vdots   \\
   Pr_{q}(m_M)
\end{pmatrix}
\] 
Here,  $M = |\M|$ is the size of the message space.  If the final
message agent $\ag{q}$ can receive is deterministic and it is $m_i$,
then the final message distribution becomes a unit vector:
\[
\vec{e}_{q,i} =    
\begin{bmatrix}
  0 &
  \ldots   &
  1 &
  \ldots   &
  0
\end{bmatrix}^{T}
\]
whose $i$th entry is $1$ and other entries are zeros.

\section{The Decision Process} \label{sec:decisionProcess}

  We now turn our attention to the core of the decision process for
  assessing impact, which is based on the following definitions of
  \emph{inferred knowledge} and of the impact that inferred
  knowledge has on $\producer$.

 % \tyq{I changed Definition~\ref{def:rv-I} below hopping to make it
%    clear. It is now based on the overall message disclosure from the
 %   producer instead of having another two agents involved. Nir: Can
 %   you please help me straight the English? }
  % 
  \begin{definition}
    \label{def:rv-I}
    Given a \fram{} $\tuple{\setagents, \setcommlinks, \M, \producer,
      m}$, let $\probDisclosuredegree{\producer}{q}$ be
    the level of disclosure of the message $m$ that the producer
    $\producer$ eventually discloses to consumer $\ag{q}$.  We
    describe the amount of knowledge that $\ag{q}$ can infer from the
    message $m$ as a random variable $ \probInfer_{q} =
    I(\probDisclosuredegree{\producer}{q}) \in \mbf{Y}$ which is
    either
    \begin{itemize}
    \item a continuous random variable whose cumulative distribution
      and density function are
      $\distributionInference{q}{\probDisclosuredegree{\producer}{q}}$
      and $\densityInference{q}{\probDisclosuredegree{\producer}{q}}$
      respectively; or
    \item a discrete random variable whose distribution is
      \[
      Pr( \probInfer_q 
      \mid \probDisclosuredegree{\producer}{q}).
      \]
    \end{itemize}
    $\mbf{Y}$ is called the space of \emph{inference}.
  \end{definition}

%
  % \tyq{We don't need the following. To ascribe $0$ to mean no
  % knowledge and $1$ to mean full inferred knowledge is should not be
  % the concern of the probabilistic model. A lot more work is needed
  % to make such a formal model on an evaluation/measurement model of
  % the fullness of a space of inferred knowledge. }
  % In particular,
  % $\probInferred{\agx}{\probDisclosuredegree{\producer}{\agx}} =
  % 0$ will imply no knowledge and
  % $\probInferred{\agx}{\probDisclosuredegree{\producer}{\agx}} =
  % 1$ inferred knowledge of equal extend as the one \agentone{} has.
% % \footnote{For simplicity, we currently assume that agents share
% %     communicated information, but do not share any inferences they may
% %     make.}.
%   We write
%   $\famRvInference{\agx}{\probDisclosuredegree{\producer}{\agx}}$
%   to represent the family of r.v.'s
%   $\rvInference{\agx}{\probDisclosuredegree{\producer}{\agx}}$.

%  \tyq{I unify the concept of benefits and damages into impact. Then I
%    split the original Definition of impact into three definitions:
%    the impact, impact distribution, and the impact measurement
%    (e.g. benefit, risk) hopping to make it clear below. Also I have
%    changed the definition of impact into the real domain instead of
%    $[0,1]$ which requires a normalization step to unify the
%    measurement of the impacts for the whole domain. Nir: Can you
%    please help me straight the English? }

  As we have previously discussed, the provision of information
  enables a recipient to make inferences, which have an effect, or
  impact on the information producer. We capture this impact as a
  point within an impact space $\mbf{Z}$. Since the producer does not
  have full information regarding a consumer's knowledge, we model
  impact probabilistically.
% However, the provision of information to undesirable agents
%   can also result in some sort of . In the following we
%   will consider an impact equal to 0 when the producer will experience
%   no harm at all by a consumer, while an impact equal to 1 means that
%   the producer will receive the highest damage possible (\ie{} total
%   destruction) by a specific consumer.

 % 
  \begin{definition}
    \label{def:rv-Z}
    Given a \fram{} $\tuple{\setagents, \setcommlinks, M, \producer,
      m}$, let $ \probInfer_q$ be the inference that a consumer
    $\ag{q}$ can make when the producer $\producer$ disseminates the
    message $m$ through the communication network.  We define the
    impact of the inferences that agent $\ag{q}$ on the producer
    $\producer$ as a real random variable $ \probImpact_q = Z_q(\probInfer_q) \in \mbf{Z}$ which is
    either
    \begin{itemize}
    \item a continuous random variable whose cumulative distribution
      and density function are $\distributionImpact{q}{\probInfer_q}$
      and \allowbreak $\densityImpact{q}{\probInfer_q}$ respectively,
      or
    \item a discrete random variable whose distributed is
      \[
      Pr( \probImpact_{q} \mid \probInfer_q).
      \]
    \end{itemize}
    The range $\mbf{Z}$ is called \emph{the impact space}.
  \end{definition}
  We concentrate on two types of impact, namely the positive and
  negative effects of the inferences made by the consumer on the
  producer. We respectively refer to these as the \emph{benefits} and
  \emph{risks} to the producer. Unlike standard utility theory, we do
  not, in general, assume that benefits and risks are directly
  comparable, and these therefore serve as two dimensions of the
  impact space.
  \begin{itemize}
  \item Benefit $\mbf{B}$: Let $\probBenefit \in \mbf{B}$ be the
    producer $\producer$'s evaluation of the benefit of inferences a
    consumer $\ag{q}$ can make following the receipt of a
    message. Following Definition~\ref{def:rv-Z}, we model benefit via
    either a continuous random variable $\rvBenefit{q}{\probInfer_q}$
    with cumulative distribution and density function
    $\distributionBenefit{q}{\probInfer_q}$ and \allowbreak
    $\densityBenefit{q}{\probInfer_q}$ respectively; or a discrete
    random variable whose distributed is $Pr( \probBenefit_{q} \mid
    \probInfer_q)$.
    
  \item Risk $\mbf{R}$: Let $\probRisk \in \mbf{R}$ be the producer
    $\producer$'s evaluation of the risk of the harm of inferences a
    consumer $\ag{q}$ can make following the receipt of a
    message. Following Definition~\ref{def:rv-Z}, we model risk vie
    either a continuous random variable $\rvRisk{q}{\probInfer_q}$
    with cumulative distribution and density function
    $\distributionRisk{q}{\probInfer_q}$ and \allowbreak
    $\densityRisk{q}{\probInfer_q}$ respectively; or a discrete
    random variable whose distributed is $Pr(
    \probRisk_{q} \mid \probInfer_q)$.

  \end{itemize}

  In \cite{Bisdikian2013} we show that several interesting properties holds in the case of continuous r.v.'s. However, in this paper we will extend the proposal previously discussed in the domain of discrete r.v.'s. An interested reader can find in Appendix \ref{sec:case-cont-rand}.

  The random variable $B_q$ implicitly captures an aspect of
  the producer's trust in the consumer, as it reflects the former's
  belief that the consumer will utilise the information in the manner
  it desires. Similarly, the random variable $R_q$ captures the
  notion of \emph{distrust} in the consumer, describing the belief
  that the consumer will utilise the information in a harmful
  manner. Note that when considering repeated interactions, these
  random variables will evolve as the producer gathers experience with
  various consumers. In such a situation each of them could represent either a
  prior distribution or a steady state. In the current work, we assume
  that a steady state has been reached, allowing us to ignore the
  problem of updating the distribution. When the context is clear, we drop the subscript notation in our r.v.'s.

\begin{figure}
\centering
\includegraphics[width=0.45\textwidth]{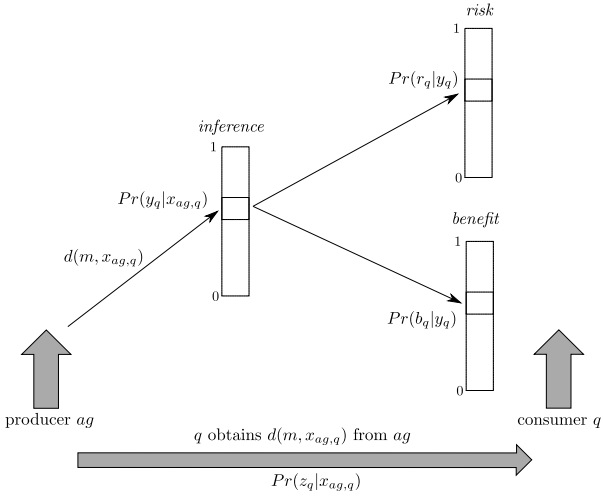}
\caption{The probabilities of inference and trust following sharing.\label{fig:trustDistribution}}
\end{figure}

  Figure \ref{fig:trustDistribution} provides a graphical
  interpretation of \emph{inference} (Definition \ref{def:rv-I}) and
  \emph{impact} (Definition \ref{def:rv-Z}), distinguishing between risk and benefit, when a producer
  \producer{} shares a message $m$ with a degree $\probDisclosuredegree{\producer}{q}$ with a
  consumer \agentq. Given $d(m, \probDisclosuredegree{\producer}{q})$, the consumer infers $y_q$
  drawn from all possible inferences. This results in an impact $z_q$
  drawn from the space of possible impacts, which is conditioned on
  the inference $y_q$ made by the consumer. This impact $z_q$ can be either a risk ($r_q$) or a benefit ($b_q$).

By assuming that the impact $z_q \in \mbf{Z}$ is independent of the
degree of disclosure $\probDisclosuredegree{\producer}{q}$ of a message $m \in \M$ given the inferred
information $y_q \in \mbf{Y}$,  we can represent the
inference distribution $ \rvInference{q}{\cdot}$ and the impact
distribution $\rvImpact{q}{\cdot}$ as conditional probability tables
in matrix notation.

The inference distribution of agent $\ag{q}$ corresponds to the
following matrix:
\[
I_{q} = 
 \begin{pmatrix}
  Pr(y_1 \mid m_1) &   Pr(y_1\mid m_2) & \cdots &   Pr(y_1\mid m_M) \\
  Pr(y_2\mid m_1) &   Pr(y_2\mid m_2) & \cdots &   Pr(y_2\mid m_M) \\
  \vdots  & \vdots  & \ddots & \vdots  \\
  Pr(y_N\mid m_1) &   Pr(y_N\mid m_2) & \cdots &   Pr(y_N\mid m_M) 
 \end{pmatrix}
\]
Each entry $Pr(y_i \mid m_j)$ represents the probability that agent
$\ag{q}$ makes inference $y_i \in \mbf{Y}$ when receiving message $m_i
\in \M$.  The $i$th column of $I_{q}$ corresponds to the inference
distribution that can be made from receiving the disclosed message
$m_i \in \M$. Note that every column will sum up to $1$ as we require that
$\Sigma_{j=1}^N Pr(b_j\mid m_i) = 1$ for a valid conditional
probability. The size of matrix $I_{q}$ is $N \times M$ where $N =
|\mbf{Y}|$ and $M = |\M|$. As $I_{q}$ is a valid conditional
probability table, there are $ (N - 1) \times M$ number of
independent parameters in $I_{q}$.

Similarly, the impact conditional probability of agent $\ag{q}$ can be
represented by an impact matrix:
\[
Z_{q} = 
 \begin{pmatrix}
  Pr(z_1 \mid y_1) &   Pr(z_1\mid y_2) & \cdots &   Pr(z_1\mid y_N) \\
  Pr(z_2\mid y_1) &   Pr(z_2\mid y_2) & \cdots &   Pr(z_2\mid y_N) \\
  \vdots  & \vdots  & \ddots & \vdots  \\
  Pr(z_K\mid y_1) &   Pr(z_K\mid y_2) & \cdots &   Pr(z_K\mid y_N) 
 \end{pmatrix}
\]
Each entry $Pr(z_k \mid y_j)$ represents the probability that agent
$\ag{q}$ cause impact $z_k \in \mbf{Z}$ to the producer $\producer$
when $\ag{q}$ can makes inference $y_j \in \mbf{Y}$.  The $j$th column
of $Z_{q}$ corresponds to the impact distribution that can occur if
the $j$th inference $y_j$ can be reached.  Again, every column will
sum up to $1$ as $\Sigma_{k=1}^K Pr(z_k \mid y_j) = 1$.  $|Z_{q}|=K
\times N$ where $N = |\mbf{Y}|$ and $K = |\mbf{Z}|$. $Z_{q}$ has $N
\times (K - 1)$ independent parameters.

\begin{corollary}
  Assume that the impact $z$ is independent of the degree of
  disclosure $x$ given the inferred information $y$.  Given
  $\vec{x}_{q}$, $I_{q}$ and $Z_{q}$, the distribution of the impact
  $\vec{\pz}_{q}$ that agent $\ag{q}$ can make to the producer
  $\producer$ is be computed by:
  \[
  \vec{\pz}_{q} = Z_{q} \times I_{q} \times \vec{x}_{q}
  \]
  where \[ \vec{\pz}_{q} =
  \begin{pmatrix}
    Pr(z_1) \\
    Pr(z_2)  \\
    \vdots   \\
    Pr(z_K)
  \end{pmatrix}
  \] whose entry $Pr(z_k)$ is the probability in which agent $\ag{q}$
  causes the $k$th impact to the producer $\producer$. $Pr(z_k)$ is
  the probability marginalizing over all possible messages and
  inferences.  $\vec{\pz}_q$ is called the \emph{the impact distribution
    vector} of a consumer $\ag{q}$.
\end{corollary}
Corresponding to the impact distribution vector $\vec{\pz}_q$, we
layout the impact evaluation into a vector $\vec{z}_q$, defined as
follows.
\[
\vec{z}_{q} =    
\begin{pmatrix}
  z_1 \\
  z_2  \\
  \vdots   \\
  z_{K}
\end{pmatrix}
\]
where $K = | \mbf{Z}|$.  Entry $z_k$ in $\vec{z}_{q}$ is the $k$-th
impact that agent $\ag{q}$ can make to the producer. $\vec{z}_q$ is
called the \emph{impact vector} by agent $\ag{q}$.
\begin{definition}
  \label{def:expected-impact}
  Given a \fram{} $\tuple{\setagents, \setcommlinks, \M, \producer,
    m}$, $\vec{\pz}_q$ (the impact probability vector) and $\vec{z}_q$
  (the impact vector) regarding agent $\ag{q}$.  The \emph{expected
    impact} regarding agent $\ag{q}$ is
  \[
  \mbb{E}\{Z_q\} = \vec{z}_q^T \times \vec{\pz}_q.
  \]
\end{definition}

Since the impact can be either a benefit or a risk, $Z_q$ can be
specialised in $Z^B_q$ (\emph{benefits probability matrix}) or $Z^R_q$
(\emph{risk probability matrix}). Correspondingly, the distribution of
impact can be either a benefit distribution or a risk distribution,
$\vec{\pb}_q$ (\emph{benefits distribution vector}) or $\vec{\pr}_q$
(\emph{risk distribution vector}); the impact vector can be either a
benefit vector $\vec{b}_q$ or a risk vector $\vec{r}_q$; the expected
impact $\mbb{E}\{z_q\}$ can either be the expected benefit
$\mbb{E}\{B_q\}$ or the expected risk $\mbb{E}\{R_q\}$.
%%%
% **Note: **
% The $k$-th impact/benefit/risk is a symbol to represent that
% impact/benefit/risk. It is not the $k$-inference. Its relationship
% to the inference is governed by a distribution which is defined by
% matrix Z, B_q, R_q (that's why originally I used Z^q, and R^q.
%%%
For notation clarity, we explicitly list benefit vector $\vec{b}_q$
and risk vector $\vec{r}_q$ respectively as follows.
\[
\vec{b}_{q} =    
\begin{pmatrix}
  b_1 \\
  b_2  \\
  \vdots   \\
  b_{K^B}
\end{pmatrix}
\]
where $K^B = | \mbf{B}|$. 
 Entry $b_k$ in $\vec{b}_{q}$ is the $k$-th benefit the
producer can obtain regarding agent $\ag{q}$. 
\[
\vec{r}_{q} =    
\begin{pmatrix}
  r_1 \\
  r_2  \\
  \vdots   \\
  r_{K^R}
\end{pmatrix}
\] where $K^R = | \mbf{R}|$.  Entry $r_k$ is the $k$-th risk the
producer is concerned with regarding agent $\ag{q}$.

We can now define the net benefit of sharing information as follows.

\begin{definition}
  \label{def:netBenefit}
  Given a \fram{} $\tuple{\setagents, \setcommlinks, \M, \producer,
    m}$, the expected benefit $ \mbb{E}\{B_q\}$ and the expected risk
  $ \mbb{E}\{R_q\}$ regarding an agent $\agent{q} \in \setagents
  \setminus \set{\producer}$, the \emph{net benefit} for the producer
  to share information with $\ag{q}$ is described by (assuming that
  the values for risk and benefit can be compared and are scaled
  appropriately for comparison):
  \[
  C_q = B_q - R_q.
  \]
  With an average, the \emph{expected net benefit} is defined as:
  \[
  \mathbb{E}\{ C_q \} = \mbb{E}\{B_q - R_q\}.
  \]
  % \[
  % \mathbb{E}\{ C_q \} = \mbb{E}\{B_q - R_q\} = \mbb{E}\{B_q\} -
  % \mbb{E}\{R_q\}\] assuming to the linearity of expectation.
\end{definition}

%NO: I don't get this next bit
%Note that the producer may \emph{value} its information $x$ according
%to the benefit it obtains. For example, for each $x$, it may value
%the information at level $\mbb{E}\{B(x)\}$ (and above) in order to
%protect itself from any expected risk $\mbb{E}\{R(x)\}$.

%%
%When \vec{x}_q is multiplied into, the messages is marginalized out.
%%
\begin{corollary}
  Assume that the impact $z$ is independent of the degree of
  disclosure $x$ given the inferred information $y$.  For agent
  $\ag{q}$, given the message disclosure distribution $\vec{x}_{q}$; 
  %which eventually reaches agent $\agx$, 
  the inference conditional
  distribution $I_{q}$;  the benefit conditional distribution
  $B_{q}$; the risk conditional distribution $R_{q}$; the
  benefit evaluation $\vec{b}_{q}$; and the risk evaluation
  $\vec{r}_{q}$. The expected benefit,  expected cost, and 
  expected net benefit that agent $\ag{q}$ can provide to the producer
  $\producer$ can be respectively computed as follows:
  \begin{align}
    \mathbb{E}\{B_q\}  & = \vec{b}_q^T \times Z^B_{q} \times I_{q} \times \vec{x}_{q} \label{eq:matrix-expect-B}\\
    \mathbb{E}\{R_q\}  & = \vec{r}^T_q \times Z^R_{q} \times I_{q} \times \vec{x}_{q} \label{eq:matrix-expect-R}\\
    \mathbb{E}\{C_q\} & = \left(
      \vec{b}^T_q \times Z^B_{q} - \vec{r}^T_q \times Z^R_{q}
    \right) \times I_{q} \times \vec{x}_{q} \label{eq:matrix-expect-C} 
  \end{align}
  Assume that there is a bijection between the spaces of benefit
  impact and cost impact, namely $B_q(y_q) = f(R_q(y_q))$ where $f$ is
  a bijection. And $B_q(y_q)$ and $f(R_q(y_q))$ have the same
  distribution after the mapping $f$ represented by the matrix $Z_q$. Then the expected net benefit
  can be simplified:
  \begin{align*}
    \mathbb{E}\{C_q\} & = \left(
      \vec{b}^T_q - \vec{r}^T_q \right) \times Z_{q} \times I_{q}
    \times \vec{x}_{q}.
  \end{align*}
\end{corollary}

\section{An Example}
\label{sec:scenario}

To illustrate our proposal, let us suppose that British Intelligence
($BI$) has  two spies, James and Alec, in place in France. James is a clever
agent, very loyal to Britain, while Alec is not as smart, and his
trustworthiness is highly questionable.  At some point, $BI$
informs the spies that in three weeks France will be invaded by a
European country: it hopes that James and Alec can recruit new agents
in France thanks to this information. However, the intelligence agency
does not specify how this invasion will take place, although they
already know it is very likely to come from the East. However, both
James and Alec are aware of the following additional pieces of
information, namely that Spain, Belgium and Italy have no interest in
invading France, while Germany does. $BI$ does not want to share the
information that the invasion will be started by Germany, because they
are the only ones aware of these plans, and a leak would result in a
loss of credibility for the UK government. Therefore, British
Intelligence has to assess the risk in order to determine whether or
not it is acceptable to inform its spies that France will be invaded
by an European country.

Formally, we can represent the above example $\fram_{BI} =
\tuple{\setagents_{BI}, \setcommlinks_{BI}, \M_{BI}, \producer_{BI},
  m_{BI}}$, where:

\begin{itemize}
\item $\set{BI, James, Alec} \subseteq \setagents_{BI}$;
\item $\set{\tuple{BI, James}, \tuple{James, BI}, \tuple{BI, Alec}, \tuple{Alec, BI}} \subseteq \setcommlinks_{BI}$;
\item $\set{m_1, m_2} \subseteq M_{BI}$ with:
  \begin{itemize}
  \item $m_1$: France will be invaded by Germany;
  \item $m_2$: France will be invaded by a European country;
  \end{itemize}
\item $\producer_{BI} = BI$;
\item $m_{BI} = m_1$;
\item $\set{James, Alec}$ are the consumers.
\end{itemize}

Suppose that $\disclosuredegree{BI}{James} =
\disclosuredegree{BI}{Alec} = x$. In other words,
$BI$ uses the same disclosure degree with both James and Alec.

In addition, $\disclosurefun(m_1, x) = m_2$, where
\begin{itemize}
\item $m_1$: France will be invaded by Germany;
\item $m_2$: France will be invaded by a European country;
\end{itemize} 

We focus on the case where $BI$ is sharing $m_{2}$ with Alec and James.

\begin{figure}[h]
  \centering
  \includegraphics[width=0.45\textwidth]{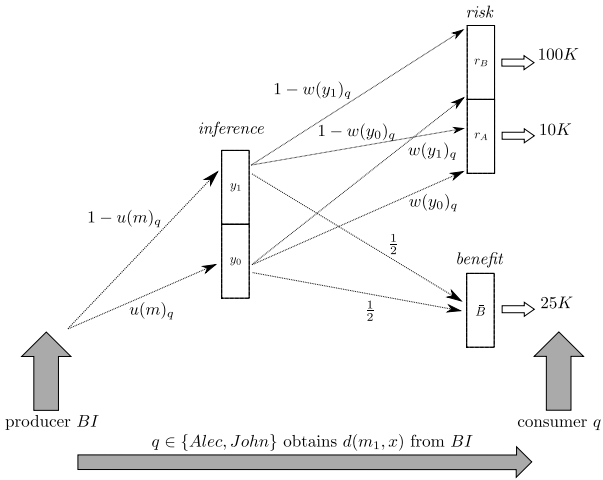}
\caption{A simple binary inference/impact case.\label{fig:impactSimplerExample}}
\end{figure}
%%%%%%%%%%%%%
To simplify the formalisation of this scenario, we will consider only
discrete random variables, allowing probability mass functions to be
used in place of densities. As illustrated in Figure
\ref{fig:impactSimplerExample}, inferences can be $y_0$ or $y_1$, with $BI$ believing that an information consumer will
make such an inference with probability $u(m_1)_q$ and $1-u(m_1)_q$
respectively if receiving message $m_1$, and with probability $u(m_2)_q$
and $1 - u(m_2)_q$ respectively if receiving message $m_2$.  Clearly, our
intention is to keep the original message ($m_{1}$) confidential,
while sharing $m_{2}$. The inference  $y_0$ is 
$m_2 = $ ``France will be invaded by a European country'', while 
$y_1$ is $m_1 = $ ``France will be invaded by
Germany''.

For each of the possible inferences there are two levels of risk,
denoted with $r_A$ and $r_B$. The risk, as well as the inferences, are independent of the agent. For simplicity, we associate a utility cost
with these two outcomes, of 10000 and 100000 respectively, as shown in
the Figure. In other words, the risk $r_A$ is $10K$ and
represents the risk of either John or Alec get captured when they are
trying to recruit new agents, while the impact $r_B$ is $100K$ and
represents the loss of credibility for the UK government due to the
sharing of information with enemy. For the sake of the example, we
will consider a fixed utility for benefit of $\bar{B} = 25K$.

With the probabilities shown in the figure, an information producing
agent is characterised by the tuple $\alpha = \linebreak \tuple{u(m_1)_q, u(m_2)_q,
 w(y_0)_q, w(1)_q}$ where $u(m_1)_q = Pr_q(y_0 \mid m_1)$ (to be read as ``the probability that agent \agentq{} will infer $y_0$ given $m$''), $u(m_2)_q =
Pr_q(y_0 \mid m_2)$, $w(y_0)_q = Pr_q( r_A \mid y_0)$ and $w(y_1)_q = Pr_q (r_A \mid
y_1)$.  Therefore, when we only concern about the result when message
$m_1$ is eventually delivered to the agents, James' behaviour can be
characterised by the tuple $\alpha_1 = \tuple{0, 0.1, 0.9, 0.9}$,
while Alec is characterised by the tuple $\alpha_2 = \tuple{0, 0.6,
  0.6, 0.4}$, which shows that if Alec infers that the invader will be
the Germany, it is more likely that he will defect, leading to the
worst possible impact for the information provider $BI$.

According to the formalism presented in this paper, we have the following space of messages: 
\[
\M = \{m_1, m_2\}
\]
Assume that $m_1$ is the final message received by the agents, we have the following vectors of eventual disclosure degree:
\[
\vec{x}_{John} = \vec{x}_{Alec} =
\begin{pmatrix}
  0 \\
  1
\end{pmatrix}
\]

Then, according to Figure \ref{fig:impactSimplerExample}, the space of
inference are composed of two possible inferences:
\[
\mbf{Y} = \{y_0, y_1\}
\]

Clearly, we have to distinguish between  John and Alec's ability to make inferences:
\[
I_{John} = 
\begin{pmatrix}
  u_{John}(1) &   u_{John}(2) \\
  1 - u_{John}(1)  &  1 - u_{John}(2)\\
\end{pmatrix}
=
\begin{pmatrix}
  0 &  0.1 \\
  1 & 0.9 \\
\end{pmatrix}
\]

\noindent
while

\[
I_{Alec} = 
\begin{pmatrix}
  u_{Alec} & u_{Alec}(2) \\
  1 - u_{Alec}  & 1 - u_{Alec}(2) \\
\end{pmatrix}
=
\begin{pmatrix}
  0 & 0.6 \\
  1 & 0.4 \\
\end{pmatrix}
\]

The risks, independent of the agent with whom information is shared is as follows:

\[
\vec{r} = 
\begin{pmatrix}
  r_A \\
  r_B
\end{pmatrix}
=
\begin{pmatrix}
  10K \\
  100K  
\end{pmatrix}
\]

\noindent

Taking the inferred messages into account, we obtain the following.

\[
R_{John} = 
 \begin{pmatrix}
  w(y_0)_{John} &   w(y_1)_{John}  \\
  1 - w(y_0)_{John} &   1 - w(y_1)_{John} 
 \end{pmatrix}
 =
\begin{pmatrix}
  0.9 & 0.9\\
  0.1 & 0.1
\end{pmatrix}
\]

\noindent
while
\[
R_{Alec} = 
 \begin{pmatrix}
  w(y_0)_{Alec} &   w(y_1)_{Alec}  \\
  1 - w(y_0)_{Alec} &   1 - w(y_1)_{Alec} 
 \end{pmatrix}
=
\begin{pmatrix}
  0.6 & 0.4\\
  0.4 & 0.6
\end{pmatrix}
\]

Moreover, since we considered a fixed benefit

\[
\vec{b} =
\begin{pmatrix}
  \bar{B}\\
\end{pmatrix}
\]

\noindent
these values depends on the inferred messages with the following distribution:

\[
B = 
\begin{pmatrix}
  \frac{1}{2} & \frac{1}{2}\\
\end{pmatrix}
\]

When agent $BI$ shares message $m$ at disclosure level $x$ with
a particular consumer agent $\agentx$, the average risk 
$\mathbb{E}\{ R_q \}$ anticipated by $BI$ is given by
\begin{equation}
  \begin{split}
    \mathbb{E}\{ R_q \} & = 
    \begin{pmatrix}
      r_A & r_B
    \end{pmatrix}
    \times
    \begin{pmatrix}
      w(y_0)_q & w(y_1)_q\\
      1 - w(y_0)_q & 1 - w(y_1)_q \\
    \end{pmatrix}
    \\
  & \quad
    \times
    \begin{pmatrix}
      u(m_1)_q & u(m_2)_q\\
      1-u(m_1)_q & 1-  u(m_2)_q\\
    \end{pmatrix}
    \times
    \begin{pmatrix}
      0 \\
      1
    \end{pmatrix}\\
% q \bigl \{ 10 w(y_0) + 100 [ 1 - w(y_0) ] \bigr \} \\
%     & \quad + (1-q) \bigl \{ 10 w(y_1)_q + 100 [ 1-w(y_1)_q ] \bigr \}
%     \\
%     & = q \bigl \{ 100 - 90 w(y_0) \bigr \} + (1 - q) \bigl [ 100 - 90
%     w(y_1)_q \bigr ]
%     \\
    & = r_B - (r_B-r_A)  \\
    & \qquad \qquad  \cdot \bigl \{ u(m_2)_q \cdot [ w(y_0)_q - w(y_1)_q ] + w(y_1)_q \bigr \} .
  \end{split}
\end{equation}

As expected, we obtain an expected risk for James of ($\mathbb{E}\{ R_{John} \} =  10K$), while for Alec, we obtain ($\mathbb{E}\{ R_{Alec} \} =  53.2K$).

%%%%%%%%%%%%%%
%\subsection{Benefit/trade-off}
%%%%%%%%%%%%%%
%In Section~\ref{sec:riskFromSharing}, we considered the risk from information sharing, however, the fact that the provider is contemplating sharing in the first place is that because he is expecting some form of benefit, \eg{} ``sell'' the piece of information for profit. To this end, let $P(x)$ represent the benefit the provider receives when shares information at level $x$; $P(x)$ could be a fixed number for each $x$, but in general we would consider it being a r.v. described by $F_P (\cdot ; x)$ and $f_P (\cdot ; x)$. When $x$ is shared, the \emph{net benefit} for the provider is described by: $C(x) = P(x) - R(x)$, with an average $\mathbb{E}\{ C(x) \} =$ $\mathbb{E}\{P(x)\} - \mathbb{E}\{R(x)\}$. The provider may \emph{value} his information $x$ according to the benefit he receives. For example, for each $x$, he may value it at level $R(x)$ (and above) to protect himself from any expected impact he may receive.

% Now let us suppose that agent $BI$ benefits at level $B(x)$ when sharing a message with disclosure level $x$, in the figure $\bar{B}(x) = \mathbb{E}\{B(x)\} = 25K$. Then, 
Similarly, the expected net benefit $\mathbb{E}\{ C_q \}$ for the spies is as follows.
\begin{equation}
  \begin{split}
    \mathbb{E}\{ C_q \} & = 
    \left(
      \begin{pmatrix}
        \bar{B} 
      \end{pmatrix}
      \times
      \begin{pmatrix}
        \frac{1}{2} & \frac{1}{2}\\
      \end{pmatrix}
      -
      \right.\\
      &\qquad
      \left.
      \begin{pmatrix}
        r_A & r_B
      \end{pmatrix}
      \times
      \begin{pmatrix}
        w(y_0)_q & w(y_1)_q\\
        1 - w(y_0)_q & 1 - w(y_1)_q \\
      \end{pmatrix}
    \right)
    \times \\
    & \qquad
    \begin{pmatrix}
      u(m_1)_q & u(m_2)_q\\
      1-u(m_1)_q & 1 - u(m_2)_q\\
    \end{pmatrix}
    \times
    \begin{pmatrix}
      0 \\
      1
    \end{pmatrix}\\
    & = \bar{B} - r_B + (r_B - r_A) \cdot \\
    & \qquad \bigl \{ u(m_2)_q \cdot [ w(y_0)_q - w(y_1)_q
    ] + w(y_1)_q \bigr \} .
  \end{split}
\end{equation}
Since $\mathbb{E}\{ C_q \} \ge 0$ is desired, we must necessarily have:
\begin{equation}
\begin{split}
r_B - \bar{B} & \le  (r_B - r_A) \cdot \\ 
& \qquad \bigl \{  u(m_2)_q \cdot [ w(y_0)_q - w(y_1)_q ] + w(y_1)_q \bigr \} \Leftrightarrow
\\
\frac{ r_B - \bar{B} }{(r_B - r_A)} & \le  u(m_2)_q \cdot [ w(y_0)_q - w(y_1)_q ] + w(y_1)_q \Leftrightarrow
\\
\frac{ r_B - \bar{B} }{(r_B - r_A)} & \le  u(m_2)_q \cdot w(y_0)_q + (1 -  u(m_2)_q) \cdot  w(y_1)_q \le 1 .
\end{split}
\label{eq:range-profit}
\end{equation}
The right hand side of the expression above represents the probability of experiencing impacts at level $r_A$ (see Fig.~\ref{fig:impactSimplerExample}), which immediately necessitates that $\bar{B} \ge r_A$. In other words, the minimum valuation of the information should be at least as large as the minimum impact expected to occur, \cf{} Definition \ref{def:netBenefit}.

From Equation \ref{eq:range-profit}, which assesses the risk and the trust model (the tuples in this discrete case), we can see that $BI$ can share the information that France is going to be invaded with James ($\frac{75}{90} \leq 0.9 \leq 1$), but not with Alec ($\frac{75}{90} \nleq 0.52 \leq 1$).

\section{Discussion and Future Work}
\label{sec:disc-future-work}

The work described in this paper makes use of an unspecified trust model as a core input to the decision making process. Our probabilistic underpinnings are intended to be sufficiently general to enable it to be instantiated with arbitrary models, such as \cite{josang02beta,teacy06travos}. Unlike these models, our work is not intended to compute a specific trust value based on some set of interactions, but rather to decide how to use the trust value output by the models.

The use of trust within a decision making system is now a prominent research topic, see \cite{Castelfranchi2010,Urbano2013} for an overview. However, the most work in this area assumes that agents will interact with some most trusted party, as determined by the trust model. This assumption reflects the basis of trust models on action and task delegation rather than information sharing. \cite{burnett11trust} is an exception to this trend; while still considering tasks, Burnett explicitly takes into account the fact that dealing with a trusted party may be more expensive, and thus leads to a lower utility when a task has relatively low potential harmful effects. Burnett's model therefore considers both risk and reward when selecting agents for interaction. However, Burnett situated his work using utility theory, while the present work allows for a more complex impact space to be used.

Another body of work relevant to this paper revolves around information leakage. Work such as \cite{mardziel11dynamic} considers what information should be revealed to an agent given that this agent should not be able to make specific inferences. Unlike our work, \cite{mardziel11dynamic} does not consider the potential benefits associated with revealing information.

Finally, there is a broad field of research devoted to assessing risk in different contexts. 
As summarised in \cite{Wang2011}, which compares seven definitions of trust\footnote{Although not considered in \cite{Wang2011}, the definition provided in \cite{Castelfranchi2010} follows the others.}, the notion of risk is the result of some combination of uncertainty about some outcome, and a (negative) payoff for an intelligent agent and his goals.
While this definition is widely accepted (with minor distinctions), different authors have different point of view when it comes to formally define what is meant by \emph{uncertainty}. In \cite{Kaplan1981}, instead of providing a formal definition of risk, the authors introduce a scenario-based risk analysis method, considering (i) the \emph{scenario}, (ii) its \emph{likelihood}, and (iii) the \emph{consequences} of that scenario. They also introduce the notion of \emph{uncertainty} in the definition of likelihood and of  consequences. Doing so allows them to address the core problem of such models, \viz{} that complete information of all possible scenarios is required. 
The connection between risk and trust has been the subject of several studies, \eg{} \cite{Tan2002} shows a formal model based on epistemic logic for dealing with trust in electronic commerce where the risk evaluation is one of the components that contribute to the overall trust evaluation, \cite{Das2004} proposes a conceptual framework showing the strict correspondence between risk and some definition of trust, \cite{Castelfranchi2010} discusses the connection between risk and trust in delegation. However, to our knowledge our work is the first attempt to consider risk assessment in trust-based decision making about information sharing.

There are several potential avenues for future work. First, we have assumed that trust acts as an input to our decision process, and have therefore not considered the interplay between risk and trust. We therefore seek to investigate how both these quantities evolve over time. To this aim, we also will investigate the connections between the shown approach and those based on game theory like \cite{Goffman1970}, as suggested by (van der Torre, personal communication, 1st Aug 2013) during the presentation of \cite{Bisdikian2013}.
Another aspect of work we intend to examine is how the trust process affects disclosure decisions by intermediate agents with regards to the information they receive. We note that agents might not propagate information from an untrusted source onwards, as they might not believe it. Such work, together with a more fine grained representation of the agents' internal beliefs could lead to interesting behaviours such as agents lying to each other \cite{caminada09truth}. Other scenarios of interest can be easily envisaged, and they will be investigated in future work. For instance, a slightly modified version of the framework proposed in this paper can be used for determining the degree of disclosure in order to be reasonably sure that a desired part of the message will actually reach a specific agent with which we do not know how to communicate. This is the situation when an organisation tries to reach an undercover agent by sharing some information with the enemy, hoping that somehow the relevant pieces of information will eventually reach the agent. Our long term goal is to utilise our approach to identify which message to introduce so as to maximise agent utility, given a knowledge rich (but potentially incomplete or uncertain) representation of a multi-agent system.

\section{Conclusions}
\label{sec:concl-future-works}

In this paper we described a framework enabling an agent to determine how much information it should disclose to others in order to maximise its utility. This framework assumes that  any disclosure could be propagated onwards by the receiving agents, and that certain agents should not be allowed to infer some information, while it is desirable that others do make inferences from the propagated information. We showed that our framework respects certain intuitions with regards to the level of disclosure used by an agent, and also identified how much  an information provider should disclose in order to achieve some form of equilibrium with regards to its utility. Potential applications can be envisaged in strategic contexts, where pieces of information are shared across several partners which can result in the achievement of a hidden agenda. %Therefore it is of primary importance to determine the level of disclosure of the information in order to received some benefit or to contribute to the coalition, without be harmed. %NO: I don't get this last line, so left it out

To our knowledge, this work is the first to take trust and risk into account when reasoning about information sharing, and we are pursuing several exciting avenues of future work in order to make the framework more applicable to a larger class of situations.

% \subsubsection*{Acknowledgements.}
% This paper is dedicated to the memory of Chatschik Bisdikian who recently passed away.

%The authors thank the reviewers for their helpful comments.

\bibliographystyle{spbasic}      % basic style, author-year citations
\bibliography{biblio}

\appendix

\section{The Case of Continuous Random Variables}
\label{sec:case-cont-rand}

By utilising Definitions \ref{def:rv-I} and \ref{def:rv-Z} we can describe the impact of disclosing a message to the consumers on the producer $\producer$.

\begin{proposition}
  \label{prop:Rdef}
  Given a \fram{} $\tuple{\setagents, \setcommlinks, \M, \producer, m}$; a consumer $\ag{q} \in \setagents$; and the
  message $\disclosurefun(m, \probDisclosuredegree{\producer}{q})$ received by
  \agentq. Let $y_q$ be the information inferred by $\ag{q}$ according
  to the r.v. $I_q(\probDisclosuredegree{\producer}{q})$ (with probability $\approx f_{I_q}(y_q;\probDisclosuredegree{\producer}{q}) \,
  dy_q$). Then, assuming that the impact $z_q$ is independent of the
  degree of disclosure $\probDisclosuredegree{\producer}{q}$ given the inferred information $y_q$,
  $\producer$ expects an impact $z_q$ described by the r.v. $Z_q(\probDisclosuredegree{\producer}{q})$ with
  density:
  \[
  f_{Z_q}(z_q ; \probDisclosuredegree{\producer}{q}) = \int_0^1 f_{Z_q} (z_q ; y_q) \, f_{I_q} (y_q ; \probDisclosuredegree{\producer}{q} ) \, d y_q .
  \]
  \begin{proof}
    \begin{equation*}
      \begin{split}
        \label{eq:cDistribution}
        F_{Z_q}(z_q ; \probDisclosuredegree{\producer}{q}) & = \Pr \{ Z_q \leq z_q | \probDisclosuredegree{\producer}{q} \} \\
        & = \int_0^1 \Pr \{ Z_q \leq
        z_q, I_q = y_q | \probDisclosuredegree{\producer}{q} \} \, dy_q
        \\
        & = \int_0^1 \Pr \{ Z_q \leq z_q | I_q = y_q , \probDisclosuredegree{\producer}{q} \} f_{I_q}(y_q;\probDisclosuredegree{\producer}{q}) \, dy_q \\
        & = \int_0^1 F_{Z_q} (z_q ; y_q) \, f_{I_q} (y_q ; \probDisclosuredegree{\producer}{q} ) \, d y_q ,
      \end{split}
    \end{equation*}

    The density function is easily derived from the distribution \linebreak
    $F_{Z_q}(z_q ; \probDisclosuredegree{\producer}{q})$ since $f_{Z_q}(z_q ; \probDisclosuredegree{\producer}{q}) = \frac{ d }{d z_q}F_{Z_q} (z_q ; \probDisclosuredegree{\producer}{q})$.
    \qed
  \end{proof}
\end{proposition}

Moreover, any time we need a single value characterisation of a
distribution, we can exploit the same idea of descriptors of a random
variable, by introducing descriptors for trust and risk. %%%NO I don't get this, how does this give us ``the same idea''?

\begin{definition}
Let $h(\cdot)$ be a function defined on $[0,1]$, and $y \in [0,1]$ be a level of inference. We define
\begin{equation}
\label{eq:AverageTH}
t_h^{Z_q} (x) = \int_0^1 h(w) \, f_{Z_q} (w;y) \, d w ,
\end{equation}
to be the $y$-\emph{trust descriptor} induced by $h(\cdot)$.
\end{definition}

We can do the same to obtain a impact descriptor:
\begin{definition}
Let $h(\cdot)$ be a function defined on $[0,1]$, and $x \in [0,1]$ be a level of disclosure.We define
\begin{equation}
\label{eq:AverageTHr}
t_h^{Z_q} (x) = \int_0^1 h(w) \, f_{Z_q} (w;x) \, d w ,
\end{equation}
to be the $x$-\emph{impact descriptor} induced by $h(\cdot)$.
\end{definition}

Typical $h(\cdot)$ include the moment generating functions, such
as $h(k) = k, k^2$, etc., and entropy $h(k)= - ln \bigl ( \allowbreak
f_K (k) \bigr )$ for the density of some r.v. $K$. In the following we
use the expectation as the risk descriptor, leaving consideration of other possible
functions for future work.

Finally, let us  illustrate two notable properties of our model. The first one is with regards to the case where a consumer can derive the full original message, which, unsurprisingly, leads to the worst case impact.

\begin{proposition}
  When a consumer is capable of gaining maximum knowledge, then
  $f_I(y;x) = \delta(y-1)$, where $\delta(\cdot)$ is the Dirac delta function, and $F_{Z} ( z ; x ) = F_{Z} (z) \triangleq F_{Z}
  (z;1)$, i.e., the risk coincides with the 1-trust (Definition~\ref{def:rv-Z}). %%%NO: how does this relate to Defn 7?

  \begin{proof}
    By the definition of the inference r.v. $I(x)$, when $\agentx$ is believed to gain maximum knowledge then the density $f_I(y;x)$ carries all its weight at the point $y=1$ for all $x$. Hence, $f_I(y;x) = \delta(y-1)$ and it follows from the definition of the Dirac delta function, see also Prop.~\ref{prop:Rdef}
    \begin{align}
    F_{Z} ( z ; x ) &= \int_0^1 F_Z (z ; y) \, f_I (y ; x ) \, d y  \nonumber\\
    &= \int_0^1 F_Z (z ; y) \, \delta (y-1) \, d y = F_Z (z;1) .
    \end{align}
    \qed
  \end{proof}
\end{proposition}

The second property pertains to the case where agent $\producer$ shares information with more than one consumer. Such situations are typically non-homogeneous as the trust and impact levels with regards to each consumer are different. Clearly, it is beneficial to identify conditions where these impacts balance (and, hence, indicate crossover thresholds) across the multiple agents.
%Finally, when $|\destination| > 1$, then some choices have to be made about the degree of disclosure to grant each agent in $\destination$.

For two agents $\agentone, \agenttwo$ having corresponding inference and behavioural trust distributions $F_{I_j} (y;x)$ and $F_{Z_j} (z;y)$, $j \in \{1 , 2\}$, for the shared information to have similar impact, $x_1$ and $x_2$ should be selected, such that the following holds.
\begin{equation}
\begin{split}
F_{Z_1}(z ; x_1) & = F_{Z_2}(z ; x_2) \Leftrightarrow \\
\int_0^1 F_{Z_1} (z;y) \, f_{I_1} (y;x_1) \, d y & = \int_0^1 F_{Z_2}(z;y) \, f_{I_2} (y;x_2) \, d y .
\end{split}
\end{equation}
Note that the above relationship implies the r.v.s $Z_1$ and  $Z_2$ are drawn from the same distribution. Such a requirement is typically unrealistic. Therefore in general one may want to consider equalities on the average, such as, finding $x_1$ and $x_2$ satisfying the following for appropriate $g(\cdot)$ functions.
\begin{equation}
\label{eq:equalRisks}
\mathbb{E}\{ g ( Z_1 (x_1) ) \} = \mathbb{E}\{ g ( Z_2 (x_2) ) \} ,
\end{equation}

\begin{proposition}
  Given that $g(z) = z$, in order to attain the same level of impact when $\producer$ shares information with $\agentone, \agenttwo$, the degrees of disclosure $x_1$ and $x_2$ for $\agentone, \agenttwo$ respectively must satisfy the following.
  \begin{equation}
  \mathbb{E}_{I_1} \bigl \{ \mathbb{E} \{ Z_1 (x_1) | I_1 \} \bigr \} = \mathbb{E}_{I_2} \bigl \{ \mathbb{E} \{ Z_2 (x_2) | I_2 \} \bigr \} .
  \end{equation}

  \begin{proof}
   The case where $g(z) = z$ corresponds to the regular averaging operator, and~\eqref{eq:equalRisks} becomes:
\begin{align*}
\MoveEqLeft \int_0^1 \int_0^1 z f_{Z_1} (z ; y) \, f_{I_1} (y ; x_1 ) \, d y \, dz \\
& = \int_0^1 \int_0^1 z f_{Z_2} (z ; y) \, f_{I_2} (y ; x_2 ) \, d y \, dz \\
\MoveEqLeft \Leftrightarrow  \\
\MoveEqLeft \int_0^1 f_{I_1} (y ; x_1 ) \biggl [ \int_0^1 z f_{Z_1} (z ; y)  \, d z \biggr ] \, dy \\
& = \int_0^1 f_{I_2} (y ; x_2 ) \biggl [ \int_0^1 z f_{Z_2} (z ; y)  \, d z \biggr ] \, dy
\\
\MoveEqLeft \Leftrightarrow \\
\MoveEqLeft \mathbb{E}_{I_1} \bigl \{ \mathbb{E} \{ Z_1 (x_1) | I_1 \} \bigr \} \\
 & = \mathbb{E}_{I_2} \bigl \{ \mathbb{E} \{ Z_2 (x_2) | I_2 \} \bigr \} .
\end{align*}
    \qed
  \end{proof}
\end{proposition}

\end{document}